%% file: EBSSS.tex
\documentclass{article}
\usepackage{ijcai19}
\usepackage{times}
\usepackage{latexsym}
\usepackage{graphicx}
\usepackage{color}
\usepackage{multirow}
\usepackage{amsmath}
\usepackage{amssymb}
\usepackage{float}
\usepackage[ruled,vlined,linesnumbered]{algorithm2e}

\newcommand{\astar}{A*}
\newcommand{\idastar}{IDA*}
\newcommand{\inlinecite}[1]{\citeauthor{#1} \shortcite{#1}}

\newcommand{\comment}[1]{}

\title{Exponential-Binary State-Space Search}

\author{
  Nathan R. Sturtevant$^1$\and
  Malte Helmert$^2$\\
  \affiliations
  $^1$Department of Computing Science, University of Alberta,
  Edmonton, Alberta, Canada\\
  $^2$Department of Mathematics and Computer Science, University of
  Basel, Switzerland\\
  \emails
  nathanst@ualberta.ca,
  malte.helmert@unibas.ch
}

\begin{document}

\maketitle

\begin{abstract}
Iterative deepening\footnote{This paper and another independent IJCAI 2019 submission have been merged into a single paper that subsumes both of them~\cite{helmert2019ibex}. This paper is placed here only for historical context. Please only cite the subsuming paper.} search is used in applications where the best cost bound for state-space search is unknown. The iterative deepening process is used to avoid overshooting the appropriate cost bound and doing too much work as a result. However, iterative deepening search also does too much work if the cost bound grows too slowly. This paper proposes a new framework for iterative deepening search called exponential-binary state-space search. The approach interleaves exponential and binary searches to find the desired cost bound, reducing the worst-case overhead from polynomial to logarithmic. Exponential-binary search can be used with bounded depth-first search to improve the worst-case performance of IDA* and with breadth-first heuristic search to improve the worst-case performance of search with inconsistent heuristics.

\end{abstract}

\section{Introduction and Motivation}

Iterative deepening has been a common approach in state-space search when the correct depth of search is unknown. With alpha-beta pruning in games, for instance, iterative deepening is broadly used, among other reasons \cite{marsland1986review}, to ensure the best action from a completed iteration is always available when the move clock runs out.
In single-agent search, iterative deepening is used in IDA* \cite{BFID85} to find an optimal solution when the running time will be dominated by the last iteration, which assumes the underlying search tree grows exponentially. If it does not grow exponentially, a tree with $N$ nodes could require as many as $\Theta(N^2)$ node expansions if each successive iteration only expands one new node.
As a result, IDA* is known to have poor performance in problem domains with non-unit edge costs.

This paper introduces a new approach, exponential-binary (EB) state-space search, which can be used to reduce the worst-case number of node expansions. If the optimal solution has cost $C^*$, EB search reduces the worst-case running time from $O(N^2)$ to $O(N \log C^*)$ while still returning an optimal solution. It achieves this by no longer expanding states in a strict best-first order. Like iterative deepening, exponential-binary state-space search is a general technique that can be applied to a range of different problems and algorithms. In addition to reducing the worst-case overhead of IDA*, this paper also demonstrates that exponential-binary state-space search can be used in problems with inconsistent heuristics to reduce the overhead of node re-expansions.

EB search uses an $f$-cost
bound to limit the cost of any path that is searched. Additionally, it
also uses lower and upper node expansion bounds during the search to
ensure that the $f$-cost bounds increase neither too slowly
(potentially causing quadratic overhead as in \idastar) nor too
quickly (causing large overhead by overshooting the optimal solution
cost). The key idea is to use a combination of exponential and binary
searches to determine an $f$-cost bound that falls into the desired
expansion number window.

Next, we introduce the necessary theory on state-space search to establish our formal results, followed by a general description of the exponential-binary state-space search framework and its application to \idastar-style and \astar-style search. We prove that our new algorithms achieve better guarantees on the number of node expansions than previous algorithms from the literature and experimentally validate the performance claims.

\input{background}

\input{algorithm}

\input{experiments}

\section{Conclusions}

We have shown that EBSS is able to address the worst-case running time of IDA* and of search inconsistent heuristics. Further research is needed to consider (1) re-using the data structures with EBGS and (2) apply the EB approach to other problems, such as suboptimal search.

\section{Acknowledgements}

This research was enabled in part by a sabbatical grant from the University of Denver and by Compute Canada (www.computecanada.ca).

\bibliography{paper}
\bibliographystyle{named}

\end{document}

%% file: background.tex
\section{State-Space Search}
\label{section:background}

The problem we consider in this paper is heuristic state-space search
with a black-box problem representation. A \emph{state
  space} $\Theta = \langle S, A, T, c, s_0, S_\star\rangle$ has a
finite set of \emph{states} $S$, a finite set of \emph{actions} $A$,
and a set of (state) \emph{transitions} $T \subseteq S \times A \times
S$. Moreover, there is a \emph{cost function} $c: A \rightarrow \mathbb
N_0$ that associates non-negative integer \emph{costs} with
actions.\footnote{Using integer costs simplifies presentation.
  Rational numbers can be represented by raising them to a common
  denominator and then dropping the denominator, as scaling all costs
  by a constant factor has no effect on semantics. Irrational
  costs can be approximated to any desired
  finite precision and then scaled to integers.}
Finally, $s_0 \in
S$ is called the \emph{initial state} and $S_\star \subseteq S$ is the
set of \emph{goal states}.

We are concerned with \emph{optimal} state-space search, where the
objective is to find a path (sequence of transitions) of minimum cost
from $s_0$ to any goal state $s_\star \in S_\star$. The cost of a
path $\pi = \langle \langle s_0, a_1, s_1\rangle, \langle s_1, a_2,
s_2\rangle, \dots, \langle s_{n-1}, a_n, s_n\rangle\rangle$ is defined
as the sum of the action costs involved: $c(\pi) = \sum_{i=1}^n
c(a_i)$. We assume throughout the paper that a state space $\Theta =
\langle S, A, T, c, s_0, S_\star\rangle$ is given and that $C^*$ is
the cost of an optimal solution.\footnote{If no solution exists, a
  state-space search algorithm should detect and report this, but for
  simplicity we focus on the behavior for solvable tasks in the
  following.}

\paragraph*{BBHS Algorithms.}

Rather than assuming that $\Theta$ is represented as an explicit graph
(which is often not feasible due to the large number of states), we consider
\emph{black-box heuristic search} (BBHS) algorithms. BBHS algorithms
do not require a declarative state-space representation such as those
used in classical planning \cite{ghallab-et-al-2004}, which makes them
more generally applicable but also means that they can only reason
about the state space in limited ways. Specifically, a BBHS algorithm
may only access the components of a state space through the following
methods: 1) $\textit{init}()$, returning the initial state; 2)
$\textit{is\_goal}(s)$, returning a boolean result indicating if $s$
is a goal state; 3) $\textit{succ}(s)$, returning a sequence of pairs
$\langle a, s'\rangle$ representing all outgoing transitions $\langle
s, a, s'\rangle$ of state $s$; 4) $\textit{cost}(a)$, returning the
cost of action $a$; and 5) $h(s)$, computing a numerical value called
the \emph{heuristic value} for state $s$. (We use $h$ to denote both
the algorithm that computes the heuristic value for a state and the
underlying mathematical mapping from states to numbers. More on
heuristics below.)

BBHS algorithms are a similar class to \emph{DXBB} algorithms as
defined by Eckerle et al.\ \shortcite{eckerle17sufficient} except
that Eckerle et al.\ consider bidirectional search and only allow
deterministic algorithms, where our interface only allows
unidirectional search and we make no assumption of determinism.

\paragraph*{Cost Measures.}

The computational complexity of BBHS algorithms is often measured
using one of three abstract cost measures: \emph{node expansions},
\emph{generated states}, or \emph{heuristic evaluations}. The number
of node expansions is the number of calls to $\textit{succ}$; the
number of heuristic evaluations is the number of calls to $h$. When
measuring the number of generated states, a call to $\textit{init}$
counts as 1, and a call to $\textit{succ}$ counts as $N$ if the given
state has $N$ outgoing transitions.

In more fine-grained analyses, it can be useful to take the cost of
data structure operations within a search algorithm into account. When
we speak of the \emph{runtime} of a search algorithm, we count all
above-mentioned abstract operations as 1 unit of cost except
\emph{succ}, which we count as 1 unit for each returned transition,
and we assume that states can be copied, compared and inserted
into/removed from/looked up in a hash table in 1 unit of time.

\paragraph*{Search Paths, Search Nodes, Graph and Tree Search.}

We define a \emph{search path} $\pi$ as a sequence of transitions that
defines a path in the state space beginning at the initial state
$s_0$. BBHS algorithms often keep track of the states they have
explored so far via so-called \emph{search nodes} $n$, and because
BBHS algorithms can only generate states via sequences of node
expansions starting from the initial state, each such node can be
associated with a generating search path. However, the concept of
paths explored by a search algorithm can also be applied to BBHS
algorithms like \idastar\ that do not use explicit node data
structures. Given a search path $\pi$, we write $\textit{state}(\pi)$
for the state at the end of the path, i.e., the next state to be
expanded if the path is to be explored further. The cost of a search
path is traditionally denoted as $g(\pi)$, so we define $g(\pi)$ as a
synonym for $c(\pi)$.

We say that a BBHS algorithm is a \emph{graph search} if it includes
tests that two states generated by the algorithm are the same. (Such
tests can eliminate paths leading to the same state as other paths
from consideration, a technique called \emph{duplicate elimination}.)
A BBHS algorithm that does not perform such tests is called a
\emph{tree search}. The most famous BBHS graph search is
\astar\ \cite{astar}, and the most famous BBHS tree search is
\idastar\ \cite{BFID85}.

\paragraph*{f-Values and Properties of Heuristics.}

The efficiency of BBHS algorithms largely hinges on the properties of
the heuristic $h$ it uses. BBHS algorithms generally require that $h$
has certain properties in order to guarantee that only optimal
solutions are returned. A heuristic $h$ is called \emph{admissible} if
$h(s) \le h^*(s)$ for all states $s$, where $h^*(s)$ denotes the cost
of an optimal (minimum-cost) path from $s$ to the nearest goal state,
or $\infty$ if no such path exists. A heuristic $h$ is called
\emph{consistent} if $h(s) \le c(a) + h(s')$ for all state transitions
$\langle s, a, s'\rangle$. The \emph{$f$-value} of a search path $\pi$
is defined as $f(\pi) = g(\pi) + h(\textit{state}(\pi))$ and provides
an estimate of the cost of a solution that extends the path $\pi$.

\paragraph*{Runtime Bounds.}

A search path $\pi$ is called \emph{highly promising} if $f(\pi) <
C^*$ and \emph{promising} if $f(\pi) \le C^*$ and $\pi$ does not include a
goal state. A state $s$ is called highly promising if there exists a
highly promising search path leading to $s$ and promising if there
exists a promising search path leading to $s$. We write $P_*$ for the
number of highly promising paths, $P_+$ for the number of promising
paths, $S_*$ for the number of highly promising states and $S_+$ for
the number of promising states of the state space.

In the following, we only consider BBHS algorithms that are guaranteed
to produce optimal solutions without requiring any properties of $h$
other than (possibly) admissibility and consistency.

\paragraph*{Bounds for Tree Search.}

A tree search \emph{must} perform a separate node expansion for each
highly promising path, and consequently $\Omega(P_*)$ is a lower bound
on the number of node expansions for any tree search.

When equipped with an oracle that gives the optimal solution cost
$C^*$, the \idastar\ algorithm could be implemented to perform
between $P_*$ and $P_+$ node expansions. In the absence of such an
oracle, there exist infinite families of state spaces for which
\idastar\ requires $\Omega(P_+^2)$ node expansions. The key to
constructing such quadratic examples is to ensure that every round of
the algorithm only leads to a small (bounded by a constant) number of
newly explored paths compared to the previous round.
While such bad scenarios for \idastar\ are rare in unit-cost state
spaces ($c(a) = 1$ for all actions $a$), they can arise naturally when
action costs vary widely.

This shortcoming of \idastar\ has been observed before, and there have
been attempts to address it by modifying the way in which the next
$f$-cost bound of \idastar\ is computed
\cite{sarkar1991reducing,vempaty1991depth,Wah94comparisonand,burns2013iterative}.
However, to the best of our knowledge no tree search in the literature
is known to have a better than quadratic worst-case bound in terms of
$P_+$. As one of our contributions, we present a tree search
performing at most $O(P_+ \log C^*)$ node expansions.

\paragraph*{Bounds for Graph Search.}

For graph search, the situation more subtly depends on the properties
of the heuristic. Any graph search that requires an admissible and
consistent heuristic and makes no further assumptions on the
properties of the heuristic \emph{must} expand all highly promising
states, so $\Omega(S_*)$ is a lower bound on the number of node
expansions. With such a heuristic, the \astar\ algorithm always
performs between $S_*$ and $S_+$ node expansions.

If the heuristic is admissible but not consistent, there exist
infinite families of state spaces, first described by
\inlinecite{DBLP:journals/ai/Martelli77}, for which \astar\ requires
$\Omega(2^{S_+})$ node expansions, making \astar\ exponentially less
efficient than blind search in the worst case. There exist algorithms
that improve this worst-case behavior, such as the B'
algorithm by \inlinecite{mero84}, but to the best of our
knowledge no BBHS algorithm described in the literature is known to
have a worst-case bound better than $O(S_+^2)$.\footnote{An algorithm
  with an $O(S_+^{1.5})$ bound has been described
  \cite{sturtevant2008using}, but the authors noted in a personal
  communication that the algorithm is not general -- it relies on an
  assumption unstated in that paper.} As one of our contributions, we
present an algorithm performing at most $O(S_+ \log C^*)$ node
expansions.

%% file: algorithm.tex
\newcommand{\funcname}[1]{\ensuremath{\textit{#1}}}
\newcommand{\varname}[1]{\ensuremath{\textit{\textmd{#1}}}}
\newcommand{\varindex}[1]{\ensuremath{\textup{\textmd{#1}}}}

\newcommand{\BoundedSearch}{\funcname{BoundedSearch}}
\newcommand{\EBSSS}{\funcname{EBSSS}}
\newcommand{\TestFBound}{\funcname{TestFBound}}
\newcommand{\NextFBound}{\funcname{NextFBound}}

\newcommand{\searchresult}{\varname{search\_result}}
\newcommand{\expandednodes}{\varname{expanded\_nodes}}
\newcommand{\solutionfound}{\varname{solution\_found}}
\newcommand{\isincomplete}{\varname{is\_incomplete}}
\newcommand{\status}{\varname{status}}

\newcommand{\Nmin}{\ensuremath{N_{\varindex{min}}}}
\newcommand{\Nmax}{\ensuremath{N_{\varindex{max}}}}
\newcommand{\fmax}{\ensuremath{f_{\varindex{max}}}}
\newcommand{\fold}{\ensuremath{f_{\varindex{old}}}}
\newcommand{\fnew}{\ensuremath{f_{\varindex{new}}}}
\newcommand{\flow}{\ensuremath{f_{\varindex{low}}}}
\newcommand{\fhigh}{\ensuremath{f_{\varindex{high}}}}
\newcommand{\fmid}{\ensuremath{f_{\varindex{mid}}}}
\newcommand{\boundtoolow}{\text{\textup{``bound too low''}}}
\newcommand{\boundtoohigh}{\text{\textup{``bound too high''}}}
\newcommand{\boundisgood}{\text{\textup{``bound is good''}}}

\newcommand{\Emax}{\ensuremath{E_{\varindex{max}}}}
\newcommand{\Nprev}{\ensuremath{N_{\varindex{prev}}}}

\section{Exponential-Binary State-Space Search}

In this section we introduce exponential-binary state-space search
(EBSSS). We think of EBSSS as a family of algorithms because it uses
another state-space search algorithm as a component, resulting in
different flavors of EBSSS depending on which component algorithm is
chosen.

Specifically, EBSSS delegates most of it work to an algorithm that
performs a bounded state-space search, denoted by $\BoundedSearch$ in
the following. $\BoundedSearch$ must accept two parameters: an
$f$-value bound ($\fmax$) and a bound on the number of permitted node
expansions ($\Nmax$). Any search algorithm can be used for
$\BoundedSearch$ if it satisfies the following properties:
\begin{enumerate}
\item It never performs more than $\Nmax$ state expansions. If
  after $\Nmax$ state expansions the search has not yet been
  completed, $\BoundedSearch$ signals with its return value
  that the bounded search remained incomplete and returns no solution.
\item If the search was completed and $\fmax \ge C^*$,
  $\BoundedSearch$ returns an optimal solution.
\item If the search was completed and $\fmax < C^*$,
  $\BoundedSearch$ signals with its return value that no
  solution of cost at most $\fmax$ exists.
\item The return value of $\BoundedSearch$ includes the
  information how many nodes were expanded.
\end{enumerate}

\begin{algorithm}[t!]
  \begin{footnotesize}
    \SetAlgoLined
    \SetKwBlock{DefineEBSSS}{function EBSSS$(c_1, c_2, \Delta)$}{end}
    \SetKwBlock{DefineTestFBound}{function TestFBound$(\fmax, \Nmin, \Nmax)$}{end}
    \SetKwBlock{DefineNextFBound}{function NextFBound$(\fold,
      \Delta, \Nmin, \Nmax)$}{end}
    \SetKwBlock{Loop}{loop do}{end}
    \SetKw{Break}{break}
    \DefineEBSSS{
      \label{eb-main}
      $\fmax \gets h(s_0)$ \\
      \Loop{
        \label{eb-main-loop-begin}
        $\searchresult \gets
        \BoundedSearch(\fmax, \infty)$
        \label{eb-main-loop-calls-bounded-search}
        \\
        \If{\searchresult.\solutionfound}{
          extract solution and terminate \\
        }
        $\Nmin \gets
        \lceil c_1 \cdot \searchresult.\expandednodes\rceil$
        \label{eb-compute-nmin} \\
        $\Nmax \gets
        \lceil c_2 \cdot \searchresult.\expandednodes\rceil$
        \label{eb-compute-nmax} \\
        $\fmax \gets \NextFBound(\fmax, \Delta, \Nmin, \Nmax)$ \\
      } \label{eb-main-loop-end}
    } \label{eb-main-end}
    \DefineTestFBound{
      \label{eb-test-f-bound-start}
        $\searchresult \gets
        \BoundedSearch(\fmax, \Nmax)$ \\
        \uIf{\searchresult.\solutionfound}{
          extract solution and terminate
          \label{eb-test-f-bound-solution-found} \\
        }
        \uElseIf{$\searchresult.\expandednodes < \Nmin$}{
          \Return \boundtoolow \\
        }\uElseIf{$\searchresult.\isincomplete$}{
          \Return \boundtoohigh \\
        }\Else{
          \Return \boundisgood \\
        }
    } \label{eb-test-f-bound-end}
    \DefineNextFBound{
      \label{eb-next-f-bound-start}
      $\flow \gets \fold + 1$
      \label{eb-flow-assignment-1}
      \label{eb-exponential-phase-start} \\
      $\fhigh \gets \fold + \Delta$ \\
      \Loop{
        $\status \gets \TestFBound(\fhigh, \Nmin, \Nmax)$ \\
        \uIf{$\status = \boundisgood$}{
          \Return $\fhigh$
          \label{eb-exponential-phase-good}
          \\
        }\ElseIf{$\status = \boundtoohigh$}{
          \Break
          \label{eb-exponential-phase-break}
        }
        $\Delta \gets 2 \cdot \Delta$ \\
        $\flow \gets \fhigh + 1$
        \label{eb-flow-assignment-2} \\
        $\fhigh \gets \fold + \Delta$ \\
      } \label{eb-exponential-phase-end}
      \While{$\flow \neq \fhigh$ \label{eb-binary-phase-start}}{
        $\fmid \gets \lfloor (\flow + \fhigh) / 2\rfloor$ \\
        $\status \gets \TestFBound(\fmid,
        \Nmin, \Nmax)$ \\
        \uIf{$\status = \boundtoolow$}{
          $\flow \gets \fmid + 1$
          \label{eb-flow-assignment-3} \\
        }\uElseIf{$\status = \boundtoohigh$}{
          $\fhigh \gets \fmid$ \\
        }\Else{
          \Return $\fmid$
          \label{eb-binary-phase-good}
          \\
        }
      }
      \Return $\flow$
      \label{eb-binary-phase-window-collapse}
      \label{eb-binary-phase-end}
      \\
    } \label{eb-next-f-bound-end}
  \end{footnotesize}
  \caption{Exponential-Binary State-Space Search} \label{EBSearch}
\end{algorithm}

\subsection*{High-Level Description of the EBSSS Algorithm}

We now describe EBSSS in detail. Pseudo-code is shown in
Algorithm~\ref{EBSearch}. The main function $\EBSSS$
(lines~\ref{eb-main}--\ref{eb-main-end}) controls the overall search
by performing a sequence of bounded state-space searches. Each
iteration of the main loop
(line~\ref{eb-main-loop-begin}--\ref{eb-main-loop-end}) performs a
search with a given $f$-bound and unbounded (= bound $\infty$) node
expansions (line~\ref{eb-main-loop-calls-bounded-search}). This
follows the basic structure of algorithms like IDA* \cite{BFID85} and
BFIDA* \cite{DBLP:journals/ai/ZhouH06}, and from the above description of the properties
of $\BoundedSearch$ it should be clear that this results in an optimal
BBHS algorithm as long as the $f$-bounds increase from iteration to
iteration.

The distinguishing characteristic of EBSSS is in how the sequence of
$f$-bounds is determined. On the one hand, we want the number of nodes
expanded in each iteration of the main loop to grow at an exponential
rate, so that the effort for the unsuccessful early iterations can be
amortized. On the other hand, we want to avoid growing the $f$-bound
so fast that the last iteration performs an unreasonably large amount
of wasted work considering $f$ values larger than $C^*$.

After an iteration that expanded $N$ nodes, we aim to
set the next $f$-bound in such a way that the number of expansions of
the next bounded search in the main loop falls into a ``target
interval'' of at least $\Nmin = \lceil c_1 \cdot N\rceil$ and at most
$\Nmax = \lceil c_2 \cdot N\rceil$ expansions
(lines~\ref{eb-compute-nmin}--\ref{eb-compute-nmax}). Here, $c_1$ and
$c_2$ are real-valued parameters that must satisfy $c_2 \ge c_1 > 1$.
For example, one might choose $c_1 = 2$ and $c_2 = 10$ to aim for at
least $2N$ and at most $10N$ expansions in the next iteration.

Finding a new $f$-bound that falls into this expansion range is not
always possible: it may be the case that no such $f$-bound exists.
Therefore, we also allow choosing an $f$-bound that results in more
than $\Nmax$ expansions, but only if we can guarantee that the
new bound is no larger than $C^*$, so that there is no wasted work
considering nodes with $f$-values larger than $C^*$.

\subsection*{Computing the New $f$-Bound}

Determining a suitable $f$-bound is the task of function $\NextFBound$
(lines~\ref{eb-next-f-bound-start}--\ref{eb-next-f-bound-end}), which
receives the current $f$-bound in parameter $\fold$, the target
expansion interval in parameters $\Nmin$ and $\Nmax$, and an integer
parameter $\Delta \ge 1$ that affects how aggressively we attempt to
increase the bound.

We say that a candidate $f$-bound is \emph{too low} if a search
with that bound would require fewer than $\Nmin$ node
expansions and \emph{too high} if it would require more than
$\Nmax$. If it is neither too low nor too high, it is
\emph{good}. The contract of $\NextFBound$ is to return a new
$f$-bound $\fnew$ that is good, or one that is too high but satisfies
$\fnew \le C^*$.

$\NextFBound$ internally uses a straight-forward function
$\TestFBound$
(lines~\ref{eb-test-f-bound-start}--\ref{eb-test-f-bound-end}) to test
if a given bound is too low, too high, or good by performing bounded
searches with an expansion limit of $\Nmax$. If these trial searches
result in a solution, we terminate the overall algorithm immediately
(line~\ref{eb-test-f-bound-solution-found}), which implies that
whenever $\TestFBound$ returns $\boundtoolow$ or $\boundisgood$ for a
candidate bound $\fmax$, we know that $\fmax < C^*$. (These are the
cases in which $\TestFBound$ completed the search for the given
$f$-bound.)

In order to compute $\fnew$, $\NextFBound$ follows a strategy from
an algorithm called \emph{exponential search} that was originally
designed to search for values in infinitely large sorted arrays
\cite{DBLP:journals/ipl/BentleyY76}. In brief, $\NextFBound$
tests the potential new $f$-bounds $\fold + 1\Delta$, $\fold +
2\Delta$, $\fold + 4\Delta$, $\fold + 8\Delta$, $\fold + 16\Delta$ and
so on until it encounters a bound that is not \emph{too low}
(lines~\ref{eb-exponential-phase-start}--\ref{eb-exponential-phase-end}).
If we encounter a bound that is \emph{good}, we can
return it directly (line~\ref{eb-exponential-phase-good}). Otherwise
we break out of the loop the first time we encounter a bound that is
too high (line~\ref{eb-exponential-phase-break}).

In this case we perform a binary search
(lines~\ref{eb-binary-phase-start}--\ref{eb-binary-phase-end}) between
the last tested bound that was too low and the first tested bound that
was too high until we either find a bound that is good
(line~\ref{eb-binary-phase-good}) or the binary search window
collapses to a single value
(line~\ref{eb-binary-phase-window-collapse}).

In the latter case, we might return a bound $\fnew$ which is not good.
If this happens, to satisfy the contract of $\NextFBound$ we must
guarantee that (1) $\fnew$ is too high (rather than too low) and (2)
$\fnew \le C^*$.

To verify (1), observe that the returned value equals the value of the
variable $\fhigh$, and the binary search part of $\NextFBound$ has the
invariant that $\fhigh$ always holds an $f$-bound
that is too high.

To verify (2), observe that the returned value also equals the value
of the variable $\flow$, and every time $\NextFBound$ assigns a value
to $\flow$, a search to the $f$-bound $\flow - 1$ has previously been
completed without finding a solution (in
line~\ref{eb-flow-assignment-1} because $\flow - 1 = \fold$, and in
lines~\ref{eb-flow-assignment-2} and~\ref{eb-flow-assignment-3}
because $\flow - 1$ was just tested by $\TestFBound$, which returned
$\boundtoolow$). This implies $\flow - 1 < C^*$ and hence $\fnew =
\flow \le C^*$ because $\flow$ and $C^*$ are integers.

\subsection*{Caching Search Results}

If implemented naively, $\EBSSS$ may perform searches that are clearly
redundant. For example, if a search triggered by $\NextFBound$ finds a
good $f$-bound, exactly the same $f$-bound will immediately be
searched again by the next iteration of the main loop of $\EBSSS$.
While such redundancies do not affect the following worst-case
performance results, they are clearly undesirable in an efficient
implementation.

These inefficiencies can be addressed without changing the pseudo-code
of the algorithm by making the component algorithm $\BoundedSearch$
remember the arguments and results of its last invocation and return
the cached result if it can detect that the result for the current
invocation must be the same. If $\BoundedSearch$ keeps track of the
lowest pruned $f$-values and highest expanded $f$-values encountered
during its search, the cached result can also be used in cases where
the $f$-bound only changes so slightly between searches that the set
of expanded nodes must necessarily be the same. Our implementation
is able to avoid such redundancies.

\subsection*{Computational Complexity}

We now turn to the computational complexity of $\EBSSS$. The
parameters $c_1$, $c_2$ and $\Delta$ can be chosen arbitrarily, but
will be considered fixed for the purposes of the analysis. In terms of
big-$O$ complexity, it is sufficient to consider the time spent inside
$\BoundedSearch$, as every loop of the algorithm calls
$\BoundedSearch$ at least once in every iteration (either directly or
via $\TestFBound$) and otherwise only performs $O(1)$ work per
iteration.

We know that $\BoundedSearch(\fmax, \Nmax)$ performs at most $\Nmax$
node expansions, so we study the complexity of $\EBSSS$ in terms of
the overall number of node expansions. For a given implementation of
$\BoundedSearch$ on a given state space, let $\Emax$ be the maximum
number of node expansions that $\BoundedSearch(\fmax, \infty)$
performs for any bound $\fmax \le C^*$, i.e., the worst-case cost of
one call to $\BoundedSearch$ for $f$-bounds that do not overshoot
$C^*$.

\paragraph*{Expansions within the Main Loop.}

We first consider the node expansions in the calls to $\BoundedSearch$
in the main loop (line~\ref{eb-main-loop-calls-bounded-search}),
ignoring the costs incurred by $\NextFBound$. We prove by induction
that each iteration performs at most $\Emax$ expansions, except
possibly the last one, which may perform up to $\lceil c_2
\Emax\rceil$ expansions. For the induction basis, the first iteration
uses a bound of $\fmax = h(s_0) \le C^*$ (we assume an admissible
heuristic) and therefore performs at most $\Emax$ expansions. For the
induction step, consider an iteration other than the first. If $\fmax
\le C^*$ in this iteration, it performs at most $\Emax$ expansions. If
$\fmax > C^*$, we must be in the final iteration. From the contract of
$\NextFBound$, the previous iteration must have found a \emph{good}
$f$-bound (otherwise it is not allowed to return $\fmax > C^*$), which
means that the number of expansions is at most $\lceil c_2
\Nprev\rceil$, where $\Nprev$ is the number of expansions in the
previous iteration. From $\Nprev \le \Emax$ (induction hypothesis), we
get a bound of $\lceil c_2 \Emax\rceil$ expansions for the last
iteration, completing the induction proof.

For the \emph{total} number of expansions of the main loop, the
contract of $\NextFBound$ guarantees that the number of expansions
grows exponentially (by a factor of at least $c_1$) between
iterations, and therefore all expansions in iterations other than the
last one disappear in big-$O$ notation, giving us an overall bound of
$O(\lceil c_2 \Emax\rceil) = O(\Emax)$ expansions in the main loop.
(Note that $c_2$ is a constant.)

\paragraph*{Expansions within $\NextFBound$.}

Consider an invocation of $\NextFBound$ in the main loop after
$\BoundedSearch(\fmax, \infty)$ expanded $N$ nodes. All calls to
$\BoundedSearch$ in this invocation use an expansion limit of $\Nmax =
\lceil c_2 N\rceil$, so the total number of expansions is bounded by
$\Nmax \cdot K$, where $K$ is the number of calls to $\TestFBound$
made by this invocation of $\NextFBound$.

The first $f$-bound tested by $\TestFBound$ is $\fold + \Delta < C^* +
\Delta$ (if we had $\fold \ge C^*$, the search would already have
terminated). All further $f$-bounds tested by $\TestFBound$ in the
exponential search phase are bounded from above by $2 C^*$ because the
$f$-bound at most doubles from one test to the next, and if it ever
reaches $C^*$, we must break out of the loop. (If the bounded search
for such a bound completes, we have found the solution and terminate.
If it does not complete, we break out of the loop because we have
found a bound that is too high.) Therefore, the highest value reached
by $\fhigh$ is bounded from above by $M = \max(C^* + \Delta, 2 C^*)$.

From the analysis of the original algorithm for exponential search in
sorted arrays \cite{DBLP:journals/ipl/BentleyY76}, it follows that $K = O(\log M) = O(\log
(\max(C^* + \Delta, 2 C^*))) = O(\log C^*)$, where we use that
$\Delta$ is a constant. Therefore, the total number of expansions of
this invocation of $\NextFBound$ is $O(K \cdot \Nmax) = O(\lceil c_2
N\rceil \cdot \log C^*) = O(N \log C^*)$.

In summary, whenever the main loop performs $N$ expansions,
$\NextFBound$ performs $O(N \log C^*$) expansions, so the total number
of expansions of the complete algorithm can be bounded by the total
number of expansions in the main loop multiplied with $O(\log C^*)$,
for a total bound of $O(\Emax \log C^*)$.

\subsection*{EBTS and EBGS}

We conclude this section by considering two concrete instances of
EBSSS: a tree search algorithm called \emph{exponential-binary tree
  search} (\emph{EBTS}) and a graph search algorithm called
\emph{exponential-binary graph search} (\emph{EBGS}).

EBTS is EBSSS where $\BoundedSearch$ is implemented like the
$f$-bounded search within IDA*, except that the $f$-bounded search
does not terminate immediately after finding solution but continues
searching for cheaper solutions, using solutions found so far for
pruning. (This modification is necessary because, unlike IDA*, EBSSS
does not increase the $f$-bound in minimal steps.) In other words,
$\BoundedSearch$ is a depth-first branch-and-bound search using $f$
for pruning.

For $f$-bounds of at most $C^*$, this implementation of
$\BoundedSearch$ never expands more than $P_+$ nodes. Therefore, we
have $\Emax \le P_+$ for EBTS, and hence our general bound for EBSSS
shows that this algorithm never performs more that $O(P_+ \log C^*)$
node expansions. As discussed in Section~\ref{section:background},
earlier tree search algorithms have worst-case expansion bounds of
$\Omega(P_+^2)$.

EBGS is EBSSS where $\BoundedSearch$ is implemented like the
$f$-bounded search in the BFIDA* algorithm
\cite{DBLP:journals/ai/ZhouH06}, except that we replace the
breadth-first search of BFIDA* with Dijkstra search. (The original
BFIDA* algorithm was only described for the unit-cost setting, but
replacing breadth-first search with Dijkstra search is the natural
generalization to general costs.)

For $f$-bounds of at most $C^*$, this implementation of
$\BoundedSearch$ never expands more than $S_+$ nodes. Therefore, we
have $\Emax \le S_+$ for EBGS, and hence our general bound for EBSSS
shows that this algorithm never performs more that $O(S_+ \log C^*)$
node expansions. As discussed in Section~\ref{section:background}, A*
requires $\Omega(2^{S_+})$ expansions in the worst case, and earlier
graph search algorithms designed to eliminate this worst case require
$\Omega(S_+^2)$ expansions.

%% file: experiments.tex
\newcommand{\Mero}{M{\'e}r\H{o}}

\section{Experimental Results}

This section evaluates the EBSS approach experimentally in three domains, the sliding tile puzzle, the pancake puzzle, and in a class of graphs due to \inlinecite{mero84} for which A* and variants such as B' require $\Omega(S_+^2)$. The code for EBSS and each of the experimental results that appear in this paper is publicly available (url redacted); each experiment can be re-run through the use of command-line parameters.

Experiments are run in a distributed manner on a research cluster which contains 2.1GHz Intel Skylake and Broadwell processors. Each job is run on a single processor. EBIDA* and IDA* were limited to 32 MB of memory per process and 24 hours of computation time. A* was limited to 64 GB of memory. Whenever A* could not solve a problem it was due to running out of memory.

To mitigate floating point rounding errors, IDA* and A* use a tolerance of $10^{-6}$ for all comparison and equality checks. The EBSS implementation uses 64-bit integers to represent heuristics and edge costs and is parameterized by the constant used to discretize costs. Using a multiple of $10^6$ gives approximately the same precision as IDA* and A*, while a multiple of $10^9$ uses higher precision.

EBSS has four parameters: the node upper bound and node lower bound, the resolution used when converting to an internal integer representation, and the $\Delta$ used for the exponential portion of the search. Each experimental result with EBSS indicates which parameters were used. In addition, we compare to an oracle that is provided with the solution cost and only needs to prove that there are no solutions with lesser cost.

\subsection{Sliding-Tile Puzzle}

The first experiment studies EBTS in the sliding-tile puzzle where the cost of moving tile $t$ is $1+\frac{1}{1+t}$. This cost function provides a variety of unique edge costs within a small range (1 to 1.5), and thus does not introduce a large number of cycles not present in the unit-cost problem. The heuristic used is Manhattan distance, where the heuristic incorporates the cost of moving the different tiles. The problem instances are the standard 100 instances \cite{BFID85}.

Experimental results are found in Table \ref{stp}. EBTS is the only algorithm that could solve all 100 problems within the resource limits. EBTS significantly outperforms IDA*, and with the best settings EBTS is within a factor of three of the oracle. A* expands fewer states than the oracle because the oracle must expand all promising paths, not just all promising states.
Using $\Delta = 1$ introduces extra overhead because the initial steps of the exponential portion of the search are too small. Using a $\Delta$ which is close to the average action cost mitigates this.

\begin{table}[tb]
\caption{IDA*, A*, and EBS on the weighted 4x4 Sliding Tile Puzzle. Expansions and generations are $\times10^6$.}
\begin{center}
    \resizebox{0.95\columnwidth}{!}{%
\begin{tabular}{| l |c|r|r|r|} \hline
Alg. & Solved & Exp. & Gen. & Time (s) \\ \hline
A* & 97 & 17.5 & 51.9 & 72.9 \\
IDA* & 57 & 62,044.3 & 185,800.5 & 14,186.9 \\
\hline 

EBTS(2,5,1e6,1) &  100 & 5,401.4 & 16,066.5 & 1254.2  \\
EBTS(2,5,1e6,1e6) &  100 & 858.6 & 2,555.2 & 202.2  \\
EBTS(10,20,1e6,1) &  100 & 2,062.5 & 6,144.9 & 452.4  \\
EBTS(10,20,1e6,1e6) &  100 & 704.6 & 2,097.6 & 164.7  \\ \hline

EBTS(2,5,1e9,1) &  100 & 4,786.8 & 14,230.0 & 1,098.9  \\
EBTS(2,5,1e9,1e9) &  100 & 855.6 & 2,545.9 & 200.9  \\
EBTS(10,20,1e9,1) &  100 & 2,256.6 & 6,727.0 & 495.5  \\
EBTS(10,20,1e9,1e9) &  100 & 704.8 & 2,098.4 & 165.5  \\ \hline
Oracle &  100 & 258.1 & 768.9 & 60.5 \\ \hline
\end{tabular}
}
\end{center}
\label{stp}
\end{table}%

\subsection{Pancake Puzzle}

The second experiment studies EBTS in the pancake puzzle where the cost of flipping a stack of $f$ pancakes is $1.0 + \frac{f}{10N}$. As with the sliding-tile puzzle this cost function falls into a small range, but the costs grow by increments of 0.005 as more pancakes are flipped. The heuristic used is the GAP heuristic \cite{helmert-socs2010} without accounting for the new cost function. Experiments are run on 100 hard instances of the 20-pancake puzzle \cite{valenzano2017analysis}.

Experimental results are found in Table \ref{pancake}. A* is able to solve 99 of the problem instances without running out of memory. EBTS* and IDA* both solve all problems. 
The same trends for EBTS hold in the pancake puzzle as in the sliding-tile puzzle. Using a $\Delta$ that is too small hurts performance, but EBTS is robust to different resolutions of the cost function.

\begin{table}[tb]
\caption{IDA*, A*, and EBS on the weighted Pancake Puzzle. Expansions and generations are $\times10^6$.}
\begin{center}
    \resizebox{0.95\columnwidth}{!}{%
\begin{tabular}{| l |c|r|r|r|} \hline
Alg. & Solved & Exp. & Gen. & Time \\ \hline
A* & 99 & 1.61 & 30.52 & 50.4 \\  
IDA* & 100 & 590.73 & 11,223.95 & 1,346.8 \\  \hline
EBTS(2,5,1e6,1) &  100 & 110.3 & 2096.3 & 253.4 \\
EBTS(2,5,1e6,1e6) &  100 & 20.7 & 393.3 & 48.3 \\
EBTS(10,20,1e6,1) &  100 & 52.0 & 987.1 & 117.8 \\
EBTS(10,20,1e6,1e6) &  100 & 10.3 & 196.2 & 24.1 \\ \hline
EBTS(2,5,1e9,1) &  100 & 102.8 & 1952.3 & 240.7 \\
EBTS(2,5,1e9,1e9) &  100 & 21.0 & 398.6 & 48.7 \\
EBTS(10,20,1e9,1) &  100 & 72.6 & 1379.2 & 162.4 \\
EBTS(10,20,1e9,1e9) &  100 & 10.3 & 196.2 & 24.1 \\ \hline
Oracle &  100 & 6.2 & 118.4 & 14.3 \\ \hline
\end{tabular}
}
\end{center}
\label{pancake}
\end{table}%

\subsection{Inconsistent Heuristics}

\begin{figure}[tb]
\begin{center}
\includegraphics[width=2.4in]{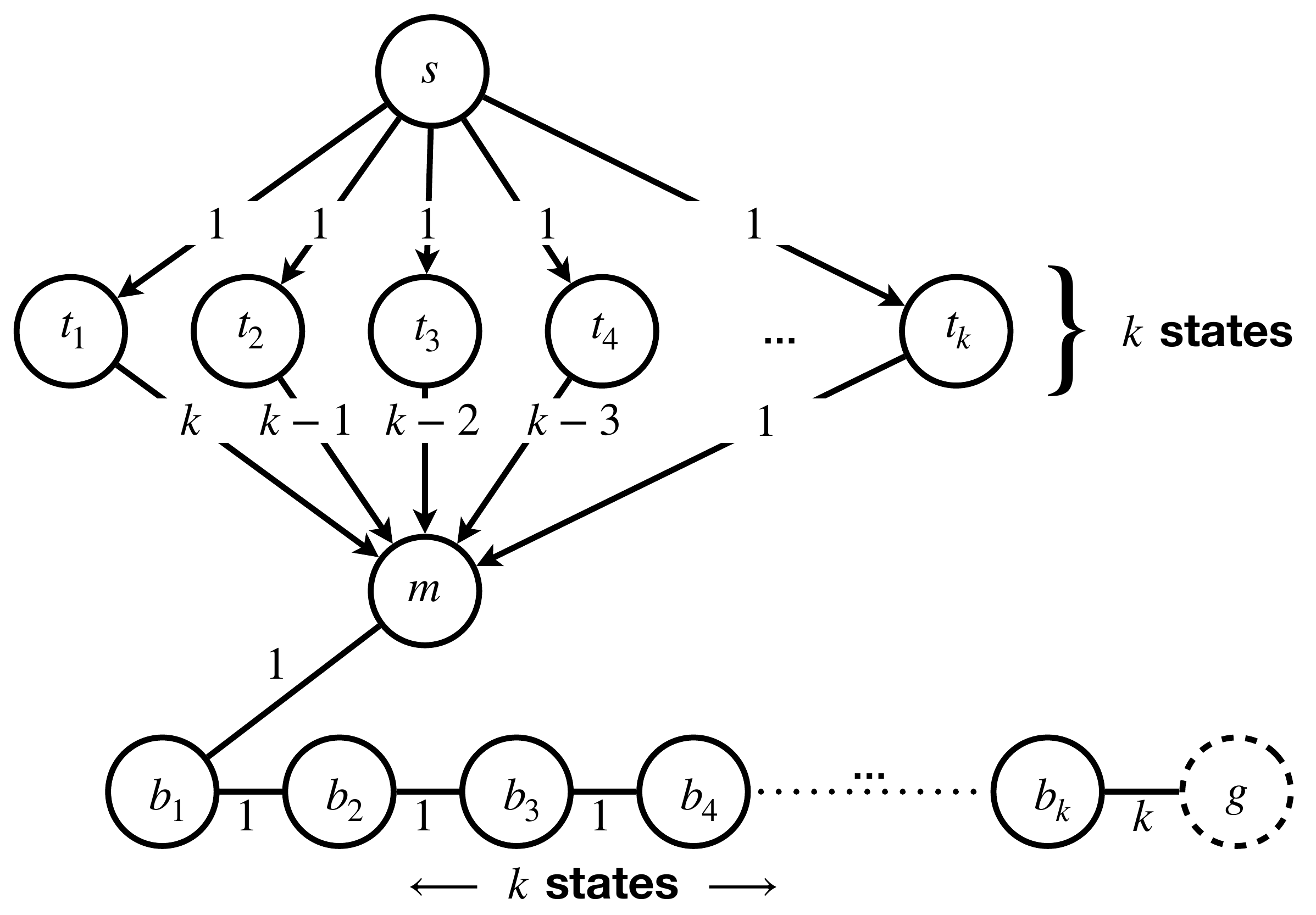}
\caption{A family of graphs from \Mero\ for which existing algorithms perform $\Omega(S_+^{2})$ expansions.}
\label{mero}
\end{center}
\end{figure}

Finally, we evaluate EBGS in problems with inconsistent heuristics. This experiment uses 
a graph, shown in Figure \ref{mero}, adapted from \inlinecite{mero84}. This graph is parameterized by $k$ and has $2k+2$ states. All states have a heuristic of 0 except each state $t_i$ which has heuristic $k+i$. Experiments are run with $k=100, 1000, 10000$. Because all edge costs have integer values, we only specify the node expansion window and the initial $\Delta$ used in the exponential search.
A*, B \cite{DBLP:journals/ai/Martelli77}, and B' \cite{mero84} all are expected to perform $O(S_+^{2})$ expansions on this graph. 
 The results of the experimental results are found in Table \ref{exp:mero}. While A* shows polynomial growth, EBGS has much better performance.

\begin{table}[tb]
\caption{Results with inconsistent heuristics. }
\small
\begin{center}
    \resizebox{0.75\columnwidth}{!}{%
\begin{tabular}{|c|l|r|r|r|}\hline
Prob. Size & Alg. & Exp. & Time (s) \\ \hline
100 & A* & 7,652 & 0.0 \\
100 & EBGS(2, 5, 1) & 2,230 & 0.0 \\
100 & EBGS(10, 20, 3) & 1,402 & 0.0 \\
100 & Oracle & 202 & 0.0 \\ \hline
1000 & A* & 751,502 & 1.3 \\
1000 & EBGS(2, 5, 1) & 32,911& 0.1 \\
1000 & EBGS(10, 20, 3) & 20,920 & 0.1 \\
1000 & Oracle & 2,002 & 0.0 \\ \hline
10000 & A* & 75,015,002 & 1,646.9 \\
10000 & EBGS(2, 5, 1) & 377,112 & 33.81s\\
10000 & EBGS(10, 20, 3) & 91,625 & 11.71s  \\
10000 & Oracle & 20,002 & 0.36 \\ \hline
\end{tabular}
}
\end{center}
\label{exp:mero}
\end{table}%

%% file: EBSSS.bbl
\begin{thebibliography}{}

\bibitem[\protect\citeauthoryear{Bentley and
  Yao}{1976}]{DBLP:journals/ipl/BentleyY76}
Jon~Louis Bentley and Andrew~Chi{-}Chih Yao.
\newblock An almost optimal algorithm for unbounded searching.
\newblock {\em Information Processing Letters}, 5(3):82--87, 1976.

\bibitem[\protect\citeauthoryear{Burns and Ruml}{2013}]{burns2013iterative}
Ethan Burns and Wheeler Ruml.
\newblock Iterative-deepening search with on-line tree size prediction.
\newblock {\em Annals of Mathematics and Artificial Intelligence},
  69(2):183--205, 2013.

\bibitem[\protect\citeauthoryear{Eckerle \bgroup \em et al.\egroup
  }{2017}]{eckerle17sufficient}
J{\"u}rgen Eckerle, Jingwei Chen, Nathan Sturtevant, Sandra Zilles, and Robert
  Holte.
\newblock Sufficient conditions for node expansion in bidirectional heuristic
  search.
\newblock In {\em Twenty-Seventh International Conference on Automated Planning
  and Scheduling (ICAPS)}, pages 79--87, 2017.

\bibitem[\protect\citeauthoryear{Ghallab \bgroup \em et al.\egroup
  }{2004}]{ghallab-et-al-2004}
Malik Ghallab, Dana Nau, and Paolo Traverso.
\newblock {\em Automated Planning: {Theory} and Practice}.
\newblock Morgan Kaufmann, 2004.

\bibitem[\protect\citeauthoryear{Hart \bgroup \em et al.\egroup }{1968}]{astar}
Peter~E. Hart, Nils~J. Nilsson, and Bertram Raphael.
\newblock A formal basis for the heuristic determination of minimum cost paths.
\newblock {\em IEEE Transactions on Systems Science and Cybernetics},
  4:100--107, 1968.

\bibitem[\protect\citeauthoryear{Helmert \bgroup \em et al.\egroup
  }{2019}]{helmert2019ibex}
Malte Helmert, Tor Lattimore, Levi H.~S. Lelis, Laurent Orseau, and Nathan~R.
  Sturtevant.
\newblock Iterative budgeted exponential search.
\newblock In {\em IJCAI. Forthcoming}, 2019.

\bibitem[\protect\citeauthoryear{Helmert}{2010}]{helmert-socs2010}
Malte Helmert.
\newblock Landmark heuristics for the pancake problem.
\newblock In {\em Third Annual Symposium on Combinatorial Search (SoCS)}, pages
  109--110, 2010.

\bibitem[\protect\citeauthoryear{Korf}{1985}]{BFID85}
Richard~E. Korf.
\newblock Depth-first iterative-deepening: An optimal admissible tree search.
\newblock {\em Artificial Intelligence}, 27(1):97--109, 1985.

\bibitem[\protect\citeauthoryear{Marsland}{1986}]{marsland1986review}
T.~Anthony Marsland.
\newblock A review of game-tree pruning.
\newblock {\em ICGA Journal}, 9(1):3--19, 1986.

\bibitem[\protect\citeauthoryear{Martelli}{1977}]{DBLP:journals/ai/Martelli77}
Alberto Martelli.
\newblock On the complexity of admissible search algorithms.
\newblock {\em Artificial Intelligence}, 8(1):1--13, 1977.

\bibitem[\protect\citeauthoryear{M{\'e}r\H{o}}{1984}]{mero84}
L{\'a}szl{\'o} M{\'e}r\H{o}.
\newblock A heuristic search algorithm with modifiable estimate.
\newblock {\em Artificial Intelligence}, 23:13--27, 1984.

\bibitem[\protect\citeauthoryear{Sarkar \bgroup \em et al.\egroup
  }{1991}]{sarkar1991reducing}
Uttam~K. Sarkar, Partha~P. Chakrabarti, Sujoy Ghose, and S.~C. De~Sarkar.
\newblock Reducing reexpansions in iterative-deepening search by controlling
  cutoff bounds.
\newblock {\em Artificial Intelligence}, 50(2):207--221, 1991.

\bibitem[\protect\citeauthoryear{Sturtevant \bgroup \em et al.\egroup
  }{2008}]{sturtevant2008using}
Nathan~R. Sturtevant, Zhifu Zhang, Robert Holte, and Jonathan Schaeffer.
\newblock Using inconsistent heuristics on {A*} search.
\newblock In {\em AAAI-Workshop on Search Techniques in Artificial Intelligence
  and Robotics}, pages 106--113, 2008.

\bibitem[\protect\citeauthoryear{Valenzano and
  Yang}{2017}]{valenzano2017analysis}
Richard~Anthony Valenzano and Danniel~Sihui Yang.
\newblock An analysis and enhancement of the gap heuristic for the pancake
  puzzle.
\newblock In {\em Tenth Annual Symposium on Combinatorial Search (SoCS)}, 2017.

\bibitem[\protect\citeauthoryear{Vempaty \bgroup \em et al.\egroup
  }{1991}]{vempaty1991depth}
Nageshwara~Rao Vempaty, Vipin Kumar, and Richard~E. Korf.
\newblock Depth-first versus best-first search.
\newblock In {\em Ninth National Conference on Artificial Intelligence (AAAI)},
  pages 434--440, 1991.

\bibitem[\protect\citeauthoryear{Wah and Shang}{1994}]{Wah94comparisonand}
Benjamin~W. Wah and Yi~Shang.
\newblock Comparison and evaluation of a class of {IDA*} algorithms.
\newblock {\em International Journal on Artificial Intelligence Tools}, pages
  493--523, 1994.

\bibitem[\protect\citeauthoryear{Zhou and
  Hansen}{2006}]{DBLP:journals/ai/ZhouH06}
Rong Zhou and Eric~A. Hansen.
\newblock Breadth-first heuristic search.
\newblock {\em Artificial Intelligence}, 170(4--5):385--408, 2006.

\end{thebibliography}
